\def\year{2019}\relax
\documentclass[letterpaper]{article} 
\usepackage{aaai19}  
\usepackage{times}  
\usepackage{helvet}  
\usepackage{courier}  
\usepackage{url}  
\usepackage{graphicx}  
\usepackage{subcaption}
\usepackage{caption}
\usepackage{enumitem,amssymb,amsmath}
\newlist{todolist}{itemize}{2}
\setlist[todolist]{label=$\square$}

\usepackage{xcolor}

\usepackage[colorlinks]{hyperref}
\hypersetup{draft}

\begin{document}
\title{RealPoint3D: Point Cloud Generation from a Single Image with Complex Background}
\author{ \textnormal{Yan Xia\textsuperscript{1}    Yang Zhang\textsuperscript{1,2}    Dingfu Zhou\textsuperscript{3}    Xinyu Huang\textsuperscript{3}    Cheng Wang\textsuperscript{1}    Ruigang Yang\textsuperscript{3}}  \\
\small{xiayan@stu.xmu.edu.cn 1205569108@qq.com cwang@xmu.edu.cn} \\
\small{\{zhoudingfu,huangxinyu01,yangruigang\}@baidu.com} \\
\textsuperscript{1} Xiamen University  \textsuperscript{2} National University of Defense and Technology \textsuperscript{3} Baidu Research
}

\maketitle
\begin{abstract}
  3D point cloud generation by the deep neural network from a single image has been attracting more and more researchers' attention. However, recently-proposed methods (e.g., \cite{fan2017point}) require the objects be captured with relatively clean backgrounds, fixed viewpoint, while this highly limits its application in the real environment. To overcome these drawbacks, we proposed to integrate the prior 3D shape knowledge into the network to guide the 3D generation. By taking additional 3D information, the proposed network can handle the 3D object generation from a single real image captured from any viewpoint and complex background. Specifically, giving a query image, we retrieve the nearest shape model from a pre-prepared 3D model database. Then, the image together with the retrieved shape model is fed into the proposed network to generate the fine-grained 3D point cloud. The effectiveness of our proposed framework has been verified on different kinds of datasets. Experimental results show that the proposed framework achieves state-of-the-art accuracy compared to other volumetric-based and point set generation methods. Furthermore, the proposed framework works well for real images in complex backgrounds with various view angles.
\end{abstract}

\section{Introduction}\label{sec:Introduction}

\begin{figure}[ht!]
	 \begin{minipage}{0.475\textwidth}
	 	\centering
	 	\includegraphics[width=1.0\textwidth]{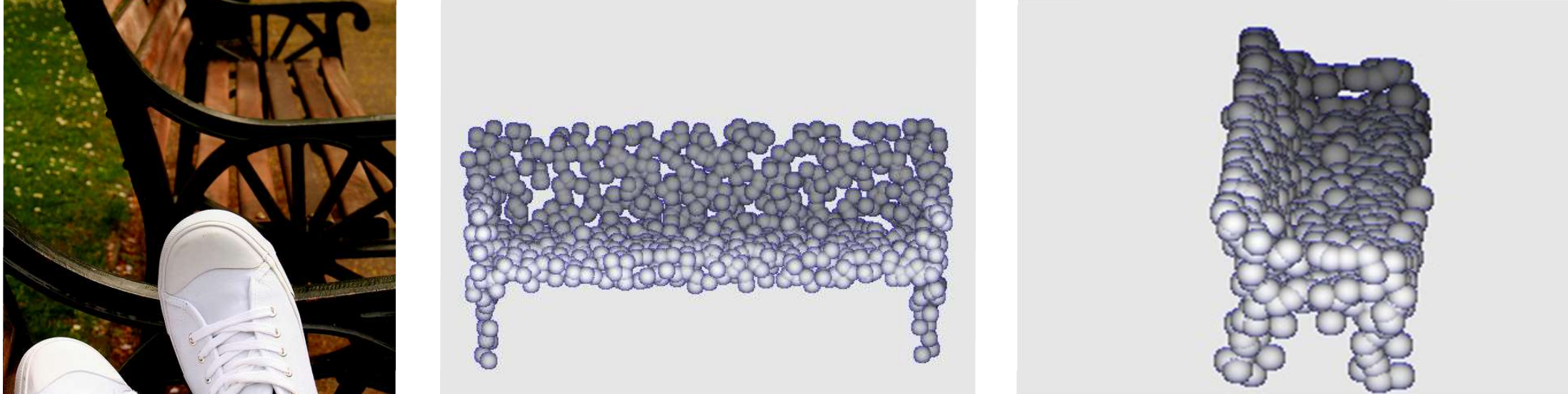}
	 	\subcaption{Generation result by the proposed RealPoint3D.}
	 	\label{subfig:1a}
	 \end{minipage}%
	 
	 \begin{minipage}{0.475\textwidth}
	 	\centering
	 		\includegraphics[width=1.0\textwidth]{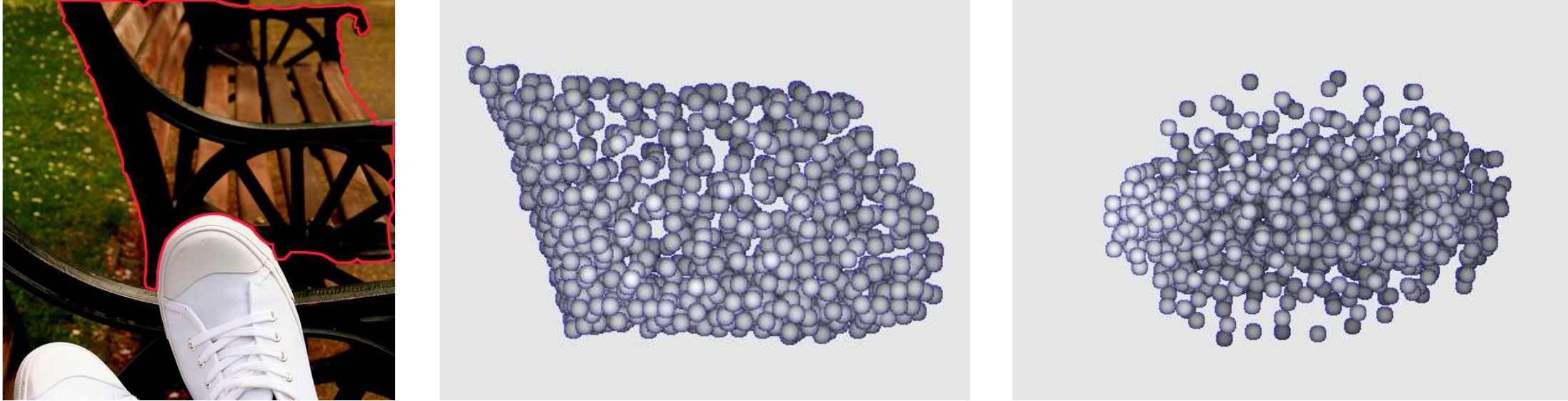}
	 	\subcaption{Generation result by PSGN \protect \cite{fan2017point}.}
	 	\label{subfig:1b}
	 \end{minipage}%
	\caption{An example of 3D point cloud generation from a real single image. Obviously, PSGN fails to generate the point cloud even the object segmentation mask is provided. The proposed method works well here without any mask information.  
	 } \label{fig:fenmian1}
\end{figure}

Recently, many learning-based methods have been proposed for depth estimation and 3D reconstruction due to the developing of deep learning. Especially, 3D reconstruction from a single image considered as an ill-posed problem, has achieved promising results by using deep neural network \cite{tatarchenko2017octree,fan2017point}. Different with traditional geometric-based approaches, learning-based methods build a complex mapping from the image space to 3D object space with the superpower representation of deep neural network. By using a single view image, the network usually need to guess the shape of object because some parts are invisible in the image. 

Before feeding into the neural network, the 3D data is usually transformed into volumetric grids or 2D images (rendered from different views) first. Then, the 2D or 3D convolution can be easily applied on the regular data. Voxelization is a common way for 3D representation and has achieved great success for object classification, detection and segmentation \cite{cicek20163d,chen2016voxresnet,meagher1980octree,wu2016learning,Choy20163D,maturana2015voxnet,tatarchenko2017octree}. Especially, the recently-proposed OGNs (Octree Generating Networks)\cite{tatarchenko2017octree} achieved impressive performances on 3D object generation task. However, the disadvantage of voxel-based method is also obvious: how to balance the sampling resolution and net efficiency is a hard problem.



To overcome this problem, some revolutionary works proposed to generate 3D point cloud for object directly in the past few years. Compared with 2D meshes or volumetric grids representations, the point cloud representation has several advantages: 1) A point cloud is a simple, uniform structure which is easy to learn by the network; 2) global geometric transformation and deformation can be easily applied to point cloud because all the points are independent and there is no connectivity information to be updated. 

Due to these advantages,  PSGN (point set generation network) \cite{fan2017point} proposed to generate 3D point cloud directly from a single image. However, this framework requires the object to be captured with a relative clean background, at a specific viewpoint and a certain distance. The reconstruction performance dropped dramatically when these requirements are satisfied (e.g., complex background). A typicality fail case for PSGN is shown in Fig. \ref{fig:fenmian1}, where the first column is the input image and the rest are generated 3D point displayed in two different views. Sub-fig \ref{subfig:1b} display the generation results by using the PSGN. Only part the bench has been generated in this case even an object mask is provided. This image is really challenge because part of the bench is occluded and some parts are truncated.  
 
In order to generate these invisible parts, we proposed to explore the prior shape information to help the 3D generation network. This knowledge has been commonly used by our human being. Specifically, we can easily imagine the 3D shape of a car even only a small part is visible and the other parts are occluded. This can succeed because we have built a huge 3D object database in our brain. By providing a small piece of object information in 2D image, we can easily find a similar 3D model in this database.

Inspired by this, we proposed a prior-knowledge-guided 3D generation framework to reconstruct the 3D point cloud from a real single image. To achieve this prior-knowledge, an on-line object retrieval stage is added before the generation framework. Here the object retrieval process is based a pre-prepared database, which stores some common 3D object models together with their corresponding image features. For each model, the image features are extracted from a number of 2D images which are rendered from different viewpoints.    

During the retrieval stage, a nearest 3D shape is searched by comparing the features extracted from the query image and these stored in the database. Compared with the 3D generation procedure, which requires dense image features, the image retrieval procedure is much robust to occlusion, changing of viewpoints and complex background, because it mainly relies on some local, sparse distinguished features. By adding the retrieved 3D model, the proposed framework can handle the 3D generation from a single image captured in the real scenario. To simplify the writing, we name the proposed network as ``RealPoint3D'' in short, where the ``Real'' consists two meanings here: on one hand, it is designed specially for image captured from the real environment; on the other hand, the output of our network is a real, complete 3D point cloud of the object including both the visible and invisible parts. 
The main contributions of this paper can be summarized as
\begin{itemize}
	\item By analyzing the drawbacks of existing methods, we proposed a prior-knowledge-guided framework to help the 3D reconstruct from a single real image;
	\item To using the prior-knowledge, we designed an end-to-end deep network RealPoint3D, which can take both the 2D image and 3D point cloud together for 3D object reconstruction. Different with the existing works, RealPoint3D can reconstruct objects with from an image with complex background and changing of viewpoints.  
	\item The RealPoint3D achieved the state-of-the-art performance on the synthetic rendered and real images, compared with volumetric and point cloud generation methods, e.g., OGN \cite{tatarchenko2017octree} and PSGN \cite{fan2017point}.
	
\end{itemize}

\section{Related Work\label{sec:Related_Work}}
\subsection{3D reconstruction from a single image}
Theoretically, the 3D structure recovery from a single projection is an ill-posed problem. To address this, many attempts were made, such as the massive SFM and SLAM \cite{Fuentes2015Visual,H2010The} methods. However, all of them require strong presumptions and abundant expertise. ShapeFromX, in which X can be the texture, specularity, shadow, etc. \cite{Barron2015Shape,Malik1997Computing,Savarese20073D}, also requires priors on natural images. 

Boosted by the large-scale dataset of 3D CAD models (e.g., ShapeNet \cite{Chang2015ShapeNet}), deep learning based generative methods have been widely employed for 3D reconstruction. Generally, they can be categorized into voxel-based and point-cloud-based methods. The 3D-GAN \cite{wu2016learning} which embedded generation task in generative adversarial nets outperforms other unsupervised learning methods with a large margin. 
In \cite{Choy20163D}, the 3D-R2N2 (3D recurrent neural network) applied the long short-term memory (LSTM) to infer 3D models by taking several images rendered from different views. By using the octree representation, OGN \cite{tatarchenko2017octree} first realized the generative method for large scene 3D reconstruction by  relieving the burden of storage and computation. 
Different with voxel-based methods, PSGN \cite{fan2017point} generates point clouds from a single image directly. However, a clean background and a fixed viewpoint are necessary conditions for good reconstruction results.  
\subsection{Shape prior guided 3D reconstruction}
Different with the natural scene, we have sufficient prior shape knowledge for artificial objects, e.g., chairs, vehicles etc. By employing this information, the 3D reconstruction becomes much easier. In \cite{su2014estimating} and \cite{huang2015single}, the authors reconstructed the depth of objects from web-site collected images by exploiting a collection of aligned 3D models of related objects shape. In \cite{karimi2015segment}, a novel shape prior formulation is proposed to split the object into multiple convex parts and then the reconstruction is formulated as a volumetric multi-label segmentation problem. 
\begin{figure*}[ht!]
	\centering
	\includegraphics[width=0.95\textwidth]{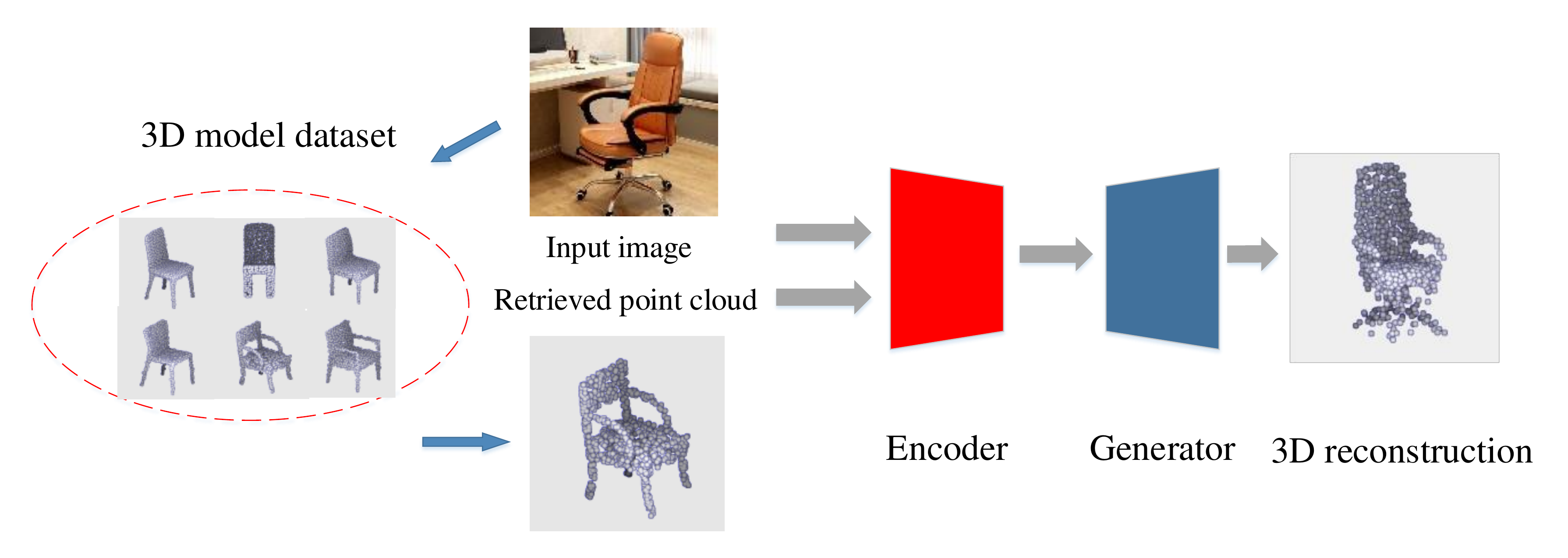}  
	\caption{The flowchart of proposed RealPoint3D framework. First of all, a nearest 3D model is retrieved from a collected 3D model database based 2D image. Then, the retrieved 3D model together with the 2D image are feed into the network for 3D reconstruction.} 
	\label{fig:overview} 
\end{figure*}

\subsection{Deep learning on point cloud}
Deep learning on point cloud attracts more and more researchers' attention recently. Voxelization, which transfers the unordered point cloud into regular grids, has been intuitively applied for 3D convolution. \cite{maturana2015voxnet} and \cite{qi2016volumetric} are two pioneers of using voxel-based methods for object detection and classification. However, they can only applied to a relatively small resolution with a sparse volume. 


PointNet \cite{qi2017pointnet} is an innovative architecture that can directly extract features from raw point cloud data which can be used for classification and segmentation tasks. PointNet++ \cite{qi2017pointnet++} which can be seen as an extension of PointNet employed multiple layers on different resolutions to increase the receptive fields of the network. Beside point clouds, many frameworks have been proposed for meshes representation, such as \cite{Bronstein2016Geometric}, spectral graph CNN \cite{Yi2017SyncSpecCNN} and Geodesic CNN (GCNN) \cite{Masci2015Geodesic}.


\section{Approach} \label{sec:Approach}

\subsection{Overview}

Different from generative-based method \cite{fan2017point}, prior shape knowledge has been taken into consideration for the 3D object reconstruction in the proposed approach. This information can help the 3D generation in following ways: 1), the prior shape can aid the network to get rid of the negative influence from clustered backgrounds; 2), the shape information can also guide the network to recover the 3D points of invisible parts which cannot been seen in the 2D image. For easy understanding, we illustrate the flowchart of RealPoint3D in Fig. \ref{fig:overview}. Generally speaking, the whole framework can be divided into two steps. First of all, deep features are extracted from the query image to retrieve the nearest 3D shape from a pre-prepared database. To robustly handle the change of viewpoint, multiple images from different viewpoints have been rendered for each model. Then, the RealPoint3D network takes the retrieval 3D model and 2D image as inputs to generate the point cloud of the object. Detailed information for each step will be introduced in the following sub-sections.



\subsection{Nearest shape retrieval}
\begin{figure*}[ht!]
	\centering
	\includegraphics[width=0.95\textwidth]{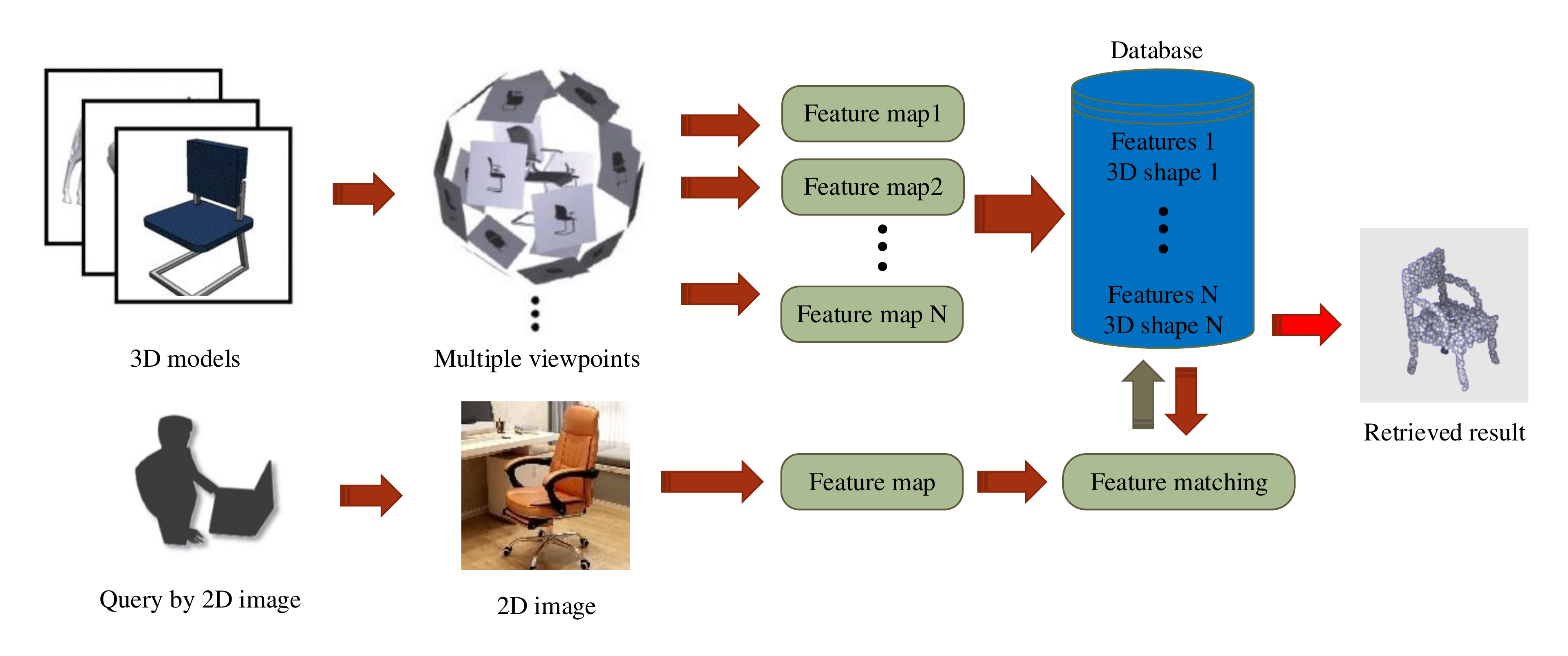}    
	\caption{Nearest shape retrieval based 2D image. First, a group of 2D images are rendered from different views for each 3D model. Then, VGG \protect \cite{simonyan2014very} network has been applied to extract deep features from these 2D images. When a query image comes, the nearest 3D model of the object related to the image can be found by comparing the 2D features extracted from the query image and the database.}
	\label{fig:retrieval_Drawing} 
\end{figure*}
 

The nearest object retrieval process is summarized in Fig. \ref{fig:retrieval_Drawing}. First of all, a database including some common used 3D models is pre-prepared in advance. For each model, multiple images are rendered from different viewpoints, e.g., 8 directions. For better simulating the real-world scenarios, background textures have been randomly added in the rendered images according to \cite{su2015render}. 

After obtaining these images, we need transform them into compact features for the further steps. The classical VGG-16 network \cite{simonyan2014very} is employed for feature computation here. A 4096-dimension feature vector is generated for each 3D model.  
Finally, a feature map database dictionary has been built for all the 3D objects. When a query image comes, the nearest object model can be obtained by comparing the features extracted from the query image with the feature map stored in database dictionary. Similar to other image retrieval methods, the similarity between two features vectors is measured by using the cosine distance, which is defined as 
\begin{equation} 
Sim(\mathbf{x},\mathbf{y}) =
\frac{\mathbf{x}\mathbf{y}}{\|\mathbf{x}\|\|\mathbf{y}\|},
\label{eq:sim}
\end{equation}
where $\mathbf{x}$ and $\mathbf{y}$ are two features to be compared.
\subsection{RealPoint3D network}\label{subsec:RealPoint3D_Network}
\begin{figure*}[ht!]
	\centering
	\includegraphics[width=1.0\textwidth]{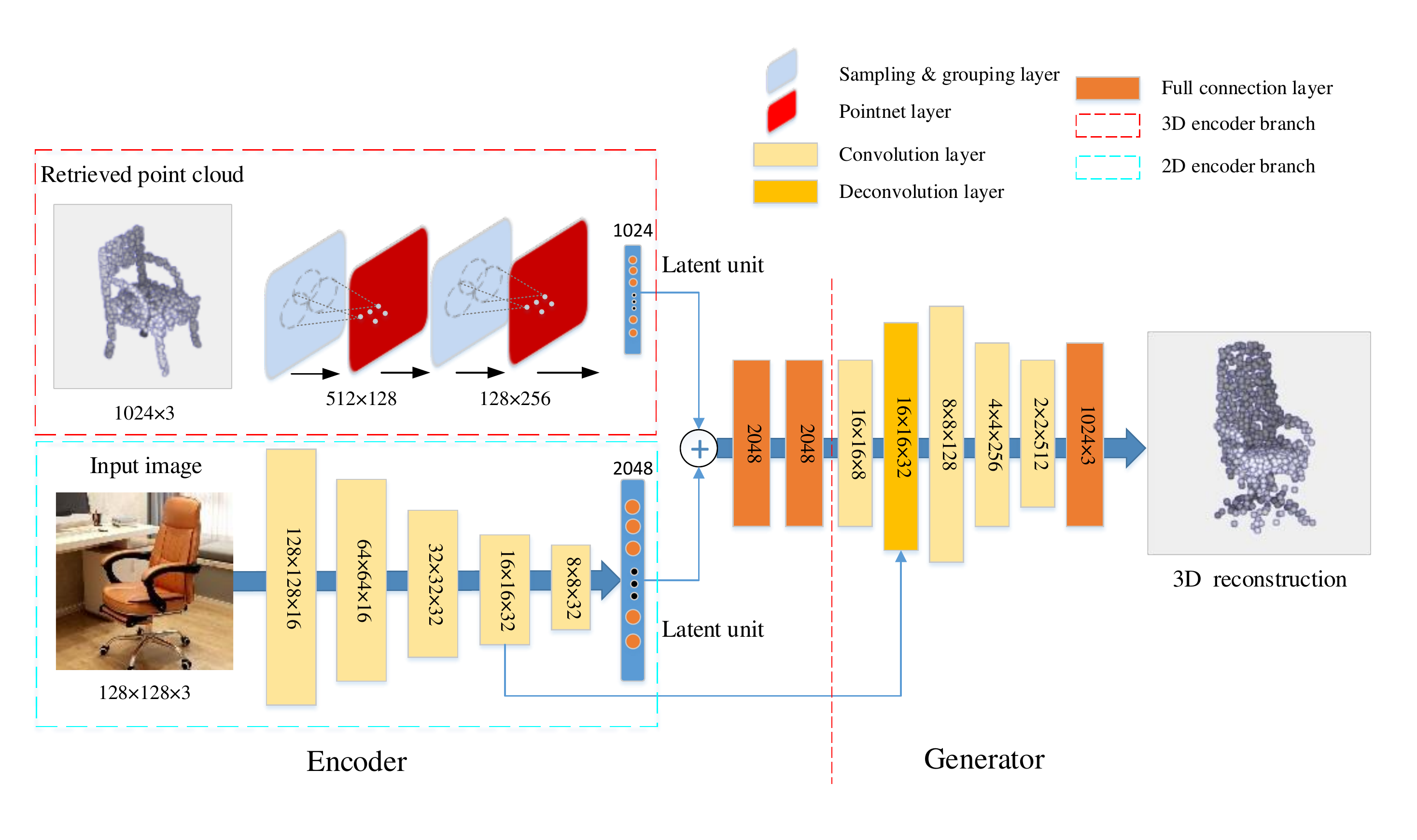}    
	\caption{A flowchart of the proposed RealPoint3D network, in which the 2D feature is extracted by an encoder network and the 3d features from point cloud are extracted similarly as Pointnet++ \protect \cite{qi2017pointnet++}. Then the two kinds of features are concatenated together for the following generator part.} 
	\label{fig:RealPoint3D_network} 
\end{figure*}

After obtaining the nearest 3D object model, $N$ (e.g., 1024) points are randomly sampled for the RealPoint3D network. An overview of the proposed network is illustrated in Fig. \ref{fig:RealPoint3D_network} for easy understanding. In the encoder part, the 2D Convolutional neural network (CNN) is used to extract the features from 2D image. Inspired by PointNet \cite{qi2017pointnet++}, Multilayer Perceptron (MLP) is employed to extract spatial information from the 3D point cloud. Then the two types of features are combined together via several fully connected layers to obtain more comprehensive features. Next, a generator stage which includes convolution and deconvolution layers is used to recover the 3D point cloud of the object. 

\subsubsection{Encoder net}
The encoder part consists two branches. One is used for 2D image and the other is for 3D point cloud. The 2D branch consists of  several convolutional and ReLU layers. Finally, the 2D part outputs a 2048-dimensional feature vector. Similar with the PointNet++, several set abstraction layers are employed in the 3D encoder branch. For each set abstraction layer, it consists of the sampling, grouping, MLP and pooling layers. In RealPoint3D, we adopt two set abstraction layers and use the multi-scale grouping strategy to obtain a global feature for the retrieved object model. Finally, the 3D encoder part outputs a 1024-dimensional feature vector. Then, the two types of features are merged together and fed to the following two fully connected layers. After obtaining the bottleneck representations, we reshape the flat feature to a 3D tensor with the size of (16,16,8).

\subsubsection{Generator part}
The generator part is made of several convolutional, deconvolutional and fully connected layers. Inspired by U-Net \cite{cicek20163d}, the low level features in the encoder stage is transferred directly to the generator stage for helping recover the details of the object. Specifically, we concatenate the feature map in the fourth convolutional layer to the deconvolutional layer. After three convolutional layers, the generator part ends with a fully connected layer with the shape of $3072$. Finally, we reshape the result to point cloud with the size of $ {1024}\times{3}$. 

\subsubsection{Influence of the 3D encoder part} To prove the effectiveness of 3D feature extraction part, a simplified version of the network has been designed for comparison. In this version, we have removed the 3D encoder part and only keep the 2D image part. The following generator part keeps the same. Experimental results proved that the performance drops dramatically due to the missing of spatial information. The comparison results of these two networks are given in subsection \ref{subsec:networkstructure}. 

\subsection{Loss function}
To enable an end-to-end training, a highly efficient and differentiable loss function must be designed. However, accurately measuring the topological similarity of two set 3D point cloud is very difficult. In addition, unlike the voxel-based approaches (e.g., 3D-R2N2), which can output $0/1$ signal to represent whether a point is inside a voxel or not. However, the proposed RealPoint3D network outputs point cloud location, therefore the Softmax function cannot be used directly here. Usually, the Hausdorff distance is selected to measure the difference between two point sets, however, this distance is sensitive to outliers. Here, we explore the Chamfer distance (CD) to measure the difference between two point sets which is defined as below

\begin{equation} 
d_{CD}=\sum_{p\in S_1}\min \limits_{q\in S_2}{\parallel p-q \parallel}_2^2 + \sum_{p\in S_2} \min \limits_{q\in S_1}{\parallel p-q \parallel}_2^2,
\label{Eq:C_distance}
\end{equation}
where $S_1, S_2\subseteq R^3$. 

In addition, CD is easily computable for two point sets and it is equal to the mean overall nearest neighbor distances. Geometrically, it induces a nice shape space. Since CD is more robust to outliers, it is a better choice for the loss function.
\subsection{Implementation and training details}
The proposed network is implemented in the framework of TensorFlow and the Adam is taken as the optimizer. To improve the performance, we choose the batch size as 32 and the gradient step as $2\times 10^5$. The learning rate automatically decays based on the number of iterations. The input image is resized to $128 \times 128 \times 3$ and the last fully connected layer produces 1024 3D points. In the encoder stage, the kernel size is set as $3 \times 3$ for all convolutional layers. In the generator stage, we set the kernel size as $5 \times 5$ for all convolutional and deconvolutional layers. In addition, the multi-scale grouping strategy is employed in the 3D point cloud encoder part. Ball query strategy is used to find neighbor points within a radius and we set $r = 0.2$ and $0.4$ for two scales. ReLU is taken as the activation function in the whole network.

\section{Experimental Results} \label{sec:Experimental_Results}
Several experiments on the rendered (e.g., ShapeNet) and real scene images (e.g., ObjectNet3D) have been considered to demonstrate the effectiveness of our proposed network. The evaluation on rendered and real images are shown in subsection \ref{subsec:rendered_image} and \ref{subsec:real_image} respectively. Finally, in order to verify the importance of prior shape model, we compared the results with and without the 3D encoder branch in subsection \ref{subsec:networkstructure}. 

\subsection{Object reconstruction on rendered images}\label{subsec:rendered_image}

\subsubsection*{Dataset:} We use the ShapeNet dataset and follow the training and testing procedure in \cite{fan2017point} and \cite{tatarchenko2017octree} for evaluation. The ShapeNet dataset is an ongoing large-scale 3D model source widely used in 3D related research fields. Our experiment is based on one of its subsets: ShapeNetCore55, which covers 55 common object categories with about $51\times10^3$ unique 3D models. As we know, it is really hard to generate the 3D ground truth models for real images. Therefore, we rendered CAD models with complex backgrounds for our training and testing. In order to test the generalization ability of our network, for each model we generate one fixed viewpoint image for training and several random viewpoints for testing. 

\subsubsection*{Competing algorithms:} We compare our approach with PSGN \cite{fan2017point} and OGN \cite{tatarchenko2017octree} in this subsection. As reported in their papers, five common categories have been selected for evaluation here. We randomly sample $1024$ points from each object for training. To have a fair comparison, we re-trained PSGN and OGN on our rendered images with complex background following their experiment settings. In addition, we used IoU (intersection over union) for evaluation with OGN and CD distance for PSGN following their corresponding papers. The results are shown in Tab. \ref{tab:results_PSGN} and \ref{tab:results_ogn} respectively.

\begin{table}[ht!]
	\begin{center}
		\begin{tabular}{|c|c|c|c|c|l|}
			\hline
			Category &PSGN&Retrieval& RP3D-v1&RP3D-v2\\
			\hline
			Sofa & 2.20 & 6.83 & 2.46 & \textbf{1.95} \\
			\hline
			Airplane & 1.00 & 3.67 & 1.38 &  \textbf{0.79} \\
			\hline
			Bench & 2.51 & 2.11 & 3.55 &  \textbf{2.11}\\
			\hline
			Car & 1.28 & 1.96 & 1.31 &  \textbf{1.26}\\
			\hline
			Chair & 2.38 & 6.91 & 2.53 &  \textbf{2.13}\\
			\hline
		\end{tabular}
	\end{center}
    \footnotesize{$^*$ We omitted the coefficient $10^{-3}$ for all the values. A smaller number represents better performance .}\\
	\caption{CD scores for different methods on complex background images, where ``Retrieval'' is the retrieved nearest shape model, ``RP3D-v1'' and ``RP3D-v2'' are our proposed methods without and with 3D point branch. We achieve lower CD in all categories. All point clouds are normalized before evaluation, and the numbers are the average point-wise distances. 
	}
	\label{tab:results_PSGN}
\end{table}

The CD scores of the testing set for the five categories are shown in Tab. \ref{tab:results_PSGN}. Our approach outperforms PSGN for all the categories, especially for the bench. The possible reason is that the retrieval accuracy of the bench is high and this information can guide the network to generate a more accurate 3D point cloud. On the contrary, PSGN obtained a similar result with us for cars. This can be explained in two aspects. On one hand, the variance of car models is not too large and this makes the 3D point cloud generation task much easy. On the other hand, the retrieval of cars is relatively difficult because there are many similar car shapes in the dataset. A relatively inaccurate retrieval shape may even mislead the generation process. Nonetheless, our generated model is more accurate. 
In addition, the selected five categories have high shape variations, which strongly indicates that the proposed approach can work well for different kinds of objects. 


For the voxel-based methods, we choose the OGN \cite{tatarchenko2017octree} for comparison here. OGN changes the original organization voxel grids to the compact octree-based structure, which significantly improves the computational efficiency and reduces the storage consumption. Five categories are evaluated here and the results are shown in Tab \ref{tab:results_ogn}. From the table, we can find that the proposed network outperforms OGN for all categories. Especially, the value has been improved from $0.046$ to $0.359$ for bench. Another interesting thing is that the proposed model can generate pretty good point cloud even the retrieved model is not good enough, e.g., car.    

\begin{table}[ht!]
\begin{center}
\begin{tabular}{|c|c|c|c|}
\hline
Category & OGN & Retrieval & RP3D-v2 \\
\hline
Sofa & 0.11 & 0.12  & \textbf{0.22} \\
\hline
Airplane & 0.15 & 0.36 &  \textbf{0.53} \\
\hline
Bench & 0.05 & 0.17  &  \textbf{0.36}\\
\hline
Car & 0.44 & 0.24 & \textbf{0.54}\\
\hline
Chair & 0.14 & 0.13 &  \textbf{0.27}\\
\hline
\end{tabular}
\end{center}
\footnotesize{$^*$A higher value represents better performance here. }\\
\caption{IoU scores for different methods on complex background images, where ``Retrieval'' is the retrieved nearest shape and ``RP3D-v2'' is our proposed methods with 3D point branch. 
}
\label{tab:results_ogn}
\end{table}


 In addition, to highlight the advantage of our method, we also designed experiments on clean background images which has been used in \cite{fan2017point} and \cite{tatarchenko2017octree}. We used the IoU as the evaluation criterion. Here, the IoU values of PSGN and OGN were picked up from their papers directly. The IoU values of sofas for PSGN and OGN are vacant in Tab. \ref{tab:results_pure_bg} because they are missing in their papers. From the table we can find that the RealPoint3D achieved the highest scores. For the car category, all the three methods give pretty good results. As we have mentioned before, the 3D generation of car is relatively easy because due to its simple structure and less shape variance.
 
 In particular, we can also find that the IOU values for the same category are different for complex and clean backgrounds in Tab. \ref{tab:results_ogn} and Tab. \ref{tab:results_pure_bg}. The former is a little lower than the latter. That demonstrates that backgrounds have strong effects on the generation process, particularly for objects with relatively complicated structures, e.g., sofa and chair. In this situation, RealPoint3D can demonstrate its strong benefit of the retrieved 3D model.

\begin{table}[ht!]
	\begin{center}
		\begin{tabular}{|c|c|c|c|c|}
			\hline
			Category & PSGN & OGN & RP3D-v2 \\
			\hline
			Sofa & - & - & \textbf{0.63} \\
			\hline
			Airplane & 0.60 & 0.59 &  \textbf{0.67} \\
			\hline
			Bench & 0.55 & 0.48 &  \textbf{0.58}\\
			\hline
			Car & \textbf{0.83} & 0.82 &  \textbf{0.83}\\
			\hline
			Chair & 0.54 & 0.48 &  \textbf{0.58}\\
			\hline
		\end{tabular}
	\end{center}
\footnotesize{$^*$A higher value represents better performance here. }\\
	\caption{IoU scores for different methods on clean background images, where ``RP3D-v2'' is our proposed methods with 3D point branch. No IoU score was reported for sofa in PSGN and OGN.
	}
	\label{tab:results_pure_bg}
\end{table}


\subsection{3D reconstruction on real images}\label{subsec:real_image}
\begin{figure}[ht!]
	\centering
	\includegraphics[width=0.475\textwidth]{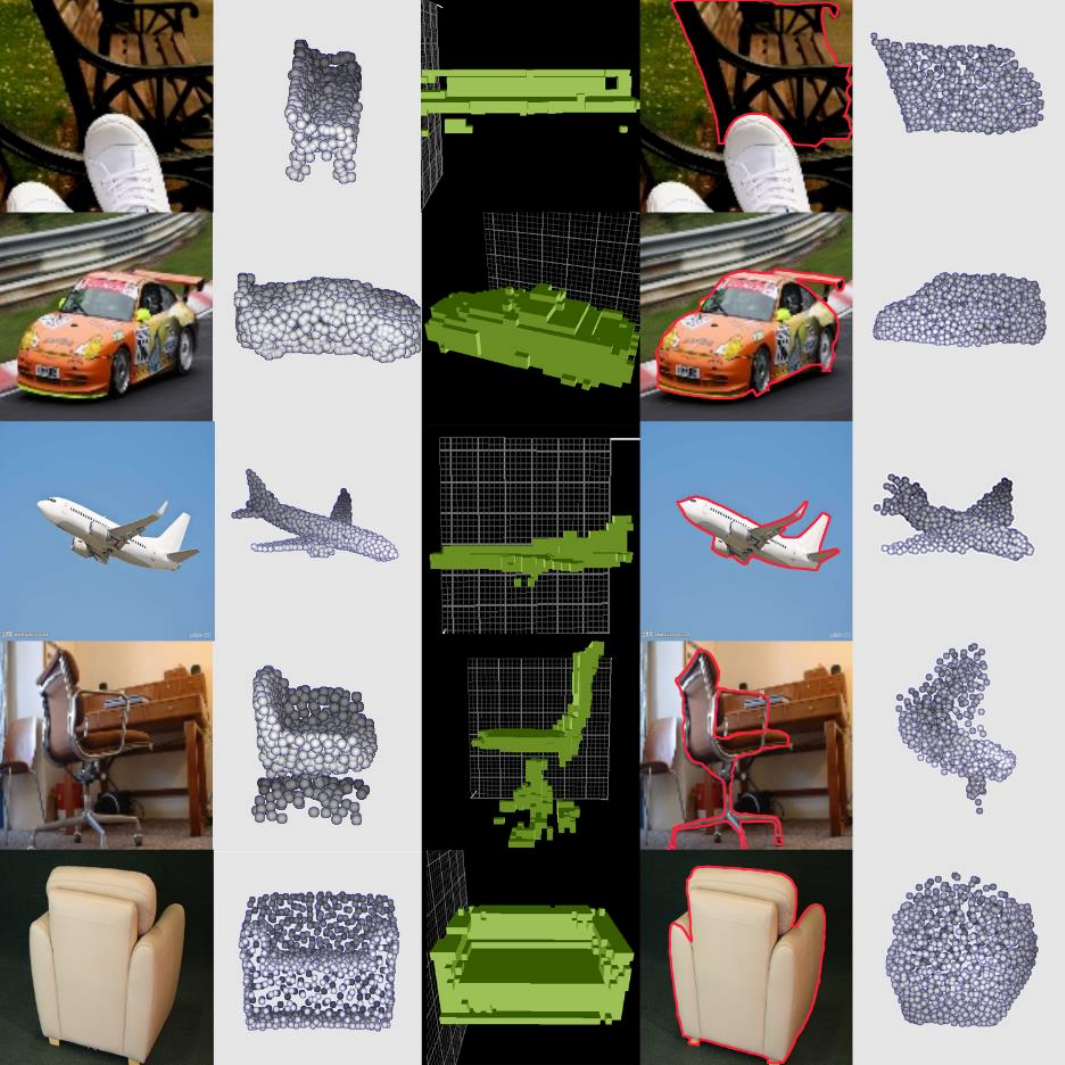}  
	\caption{3D reconstruction from real images on ObjectNet3D dataset. The orders from left to right: input 2D image, ``RP3D-v2'', PSGN with object segmentation masks, PSGN without object segmentation masks and OGN.}
	\label{fig:real_compare} 
\end{figure}

\subsubsection*{Dataset:} We use the ObjectNet3D dataset and follow the training and testing procedure used in \cite{fan2017point} and \cite{tatarchenko2017octree}. ObjectNet3D is another large-scale 3D model dataset including about 100 categories, 44147 3D shapes and 201888 objects in 90127 images. We tested our model on this dataset because it has many real world photos. 
\subsubsection*{Competing algorithms:} We compare our approach to PSGN \cite{fan2017point} and OGN \cite{tatarchenko2017octree}. Five common categories have been selected for evaluation here and some results are visualized in Fig. \ref{fig:real_compare}. From the top row to bottom, the five categories are bench, car, airplane, chair and sofa respectively. Particularly, for RealPoint3D and OGN, we take the whole image (the first column of Fig. \ref{fig:real_compare}) as the inputs for 3D generation; for PSGN, only the foreground objects inside the masks (the fourth column of Fig. \ref{fig:real_compare}) are used for generation because it will fail if we take the whole image as input.
To highlight the strength of our proposed method, five types of objects are chosen here for comparison. The second, third and the fifth columns are reconstruction results of RealPoint3D, OGN and PSGN respectively. From Fig. \ref{fig:real_compare}, we can find that RealPoint3D achieves more fine-gained object details than the other two methods. For example, at the first row of Fig. \ref{fig:real_compare}, RealPoint3D can recover the legs of bench, while the other two methods totally miss them. Even with object masks, PSGN still cannot recover the detailed structures of the bench.  

In this experiment, we can see that the proposed method is more applicable for the real word. In particular, a single image cannot provide enough information for 3D reconstruction. The geometry for invisible parts has to be guessed by the network. Therefore, PSGN and OGN can only get a global shape without fine-grained details. With the help of retrieved nearest 3D shape, the proposed network can get a better result, especially for the invisable parts.  

\subsection{Network structure comparison}\label{subsec:networkstructure}
To verify the influence of retrieved nearest 3D object, we set the following experiment to compare the reconstruction results without and with 3D encoder part. Here, RP3D-v1 represents the network without 3D encoder part and RP3D-v2 is the full proposed network.

\begin{figure}[ht!]
	\centering
	\includegraphics[width=0.475\textwidth]{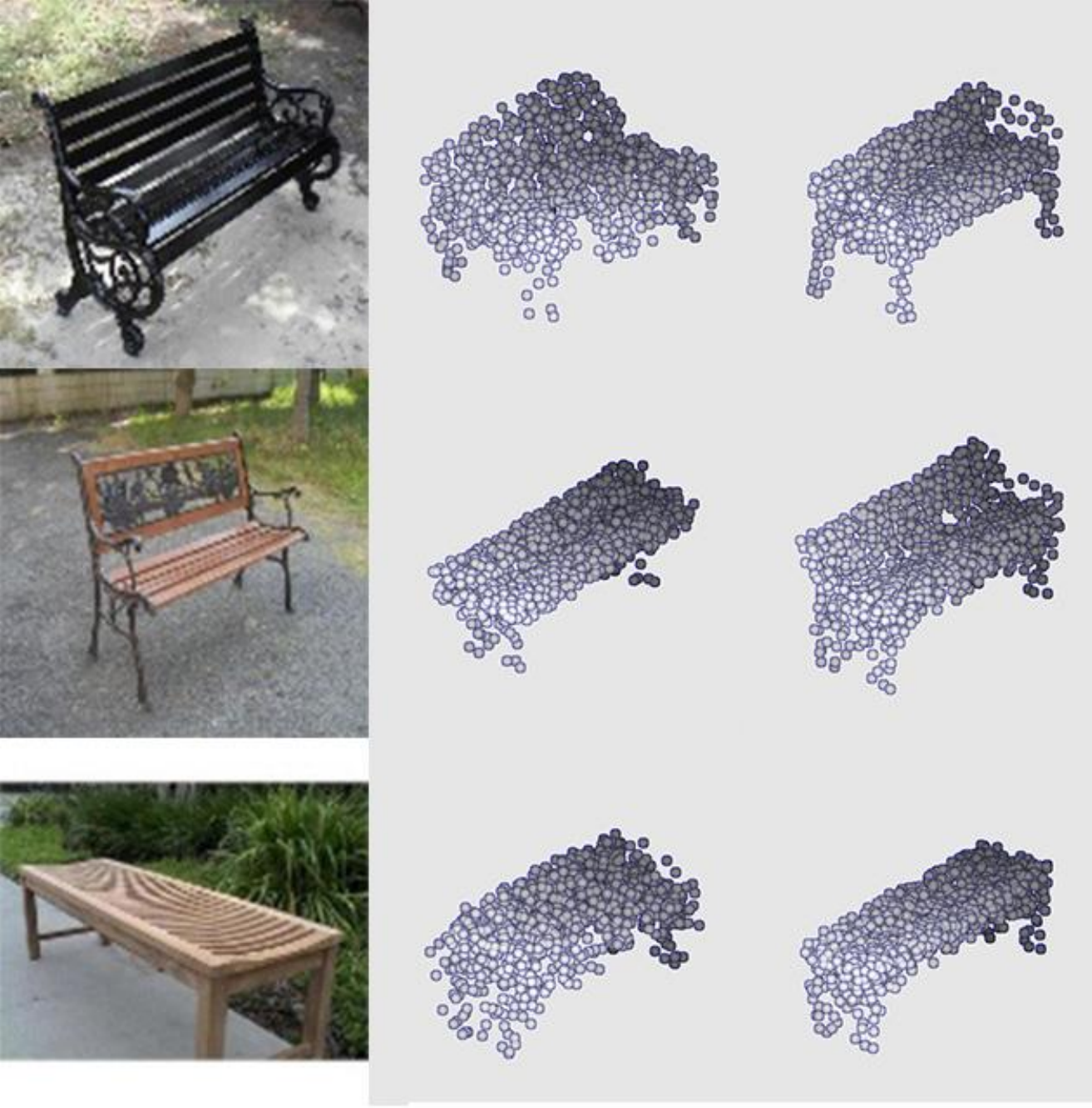}  
	\caption{Network structure comparison. Each method occupies one column, the first column is the 2D image, and the two methods are ``RP3D-v1'' and ``RP3D-v2''.} 
	\label{fig:real_car} 
\end{figure}

Some quantitative results have been shown in Tab. \ref{tab:results_PSGN}. Unsurprisingly, RP3D-v2 gives better results than RP3D-v1. We also performed some experiments on real images with different viewpoints which are displayed in Fig. \ref{fig:real_car}. The RP3D-v1 network misses the details and even generates some wrong parts of the objects. On the contrary, RP3D-v2 recovers more details of the objects.
\subsection{Time complexity}\label{sec:time_complexity}
In current implementation, 500 epochs are set during the training stage. It takes approximately 10 hours on 5 P100 GPUs. During the testing, it costs approximately 0.1s per image on a laptop with CPU. The computation efficiency is similar to PSGN but is significantly faster than OGN which takes about 1.6 s per image.


\section{Conclusion and Future Works}\label{sec:Conclusion}
In this paper, we designed a novel generation network which is more suitable for 3D fine-grained reconstruction from a single image in the real scenario. Different with the previous generative methods, a nearest 3D point clouds retrieval part is added before the main generation network. This can prompt the generator to reconstruct more object details from images with complex backgrounds and changing viewpoints. We achieved state-of-the-art performance on reconstruction from real images with complex background in comparison with other generative methods. However, the nearest object retrieval is separate part in this work. In future, we aim to integrate this part together and design an end-to-end framework.


\vspace{2cm}
\bibliographystyle{aaai}
\bibliography{bare_jrnl_V6}

\end{document}